\documentclass[conference]{IEEEtran}

\IEEEoverridecommandlockouts
\usepackage{cite}
\usepackage{amsmath,amssymb,amsfonts}
\usepackage{algorithmic}
\usepackage{graphicx}
\usepackage{tikz}
\usepackage{textcomp}
\def\BibTeX{{\rm B\kern-.05em{\sc i\kern-.025em b}\kern-.08em
    T\kern-.1667em\lower.7ex\hbox{E}\kern-.125emX}}

\begin{document}

\title{Optimal Sensor Data Fusion Architecture for Object Detection in Adverse Weather Conditions\\
}
% Institute of Measurement, Control, and Microtechnology, Ulm University, 89081 Ulm, Germany
\author{\IEEEauthorblockN{ Andreas Pfeuffer, and Klaus Dietmayer}
\IEEEauthorblockA{Institute of Measurement, Control, and Microtechnology, Ulm University, 89081 Ulm, Germany} %\\
Email: \{andreas.pfeuffer, klaus.dietmayer\}@uni-ulm.de}
%\and

\newcommand\Tstrut{\rule{0pt}{2.0ex}}         % = `top' strut
\newcommand\Bstrut{\rule[-0.9ex]{0pt}{0pt}} 
\newcommand\Mstrut{\rule[-0.0ex]{0pt}{0pt}}

\newcommand\copyrighttext{%
  \footnotesize This work has been submitted to the IEEE for possible publication. Copyright may be transferred without notice, after which this version may no longer be accessible.}%
\newcommand\copyrightnotice{%
\begin{tikzpicture}[remember picture,overlay]%
\node[anchor=south,yshift=10pt] at (current page.south) {\fbox{\parbox{\dimexpr\textwidth-2cm}{\copyrighttext}}};%
\end{tikzpicture}%
\vspace{-10pt}%
}

\maketitle
\copyrightnotice

\begin{abstract}
	A good and robust sensor data fusion in diverse weather conditions is a quite challenging task. There are several fusion architectures in the literature, e.g. the sensor data can be fused right at the beginning (Early Fusion), or they can be first processed separately and then concatenated later (Late Fusion). In this work, different fusion architectures are compared and evaluated by means of object detection tasks, in which the goal is to recognize and localize predefined objects in a stream of data. Usually, state-of-the-art object detectors based on neural networks are highly optimized for good weather conditions, since the well-known benchmarks only consist of sensor data recorded in optimal weather conditions. Therefore, the performance of these approaches decreases enormously or even fails in adverse weather conditions. In this work, different sensor fusion architectures are compared for good and adverse weather conditions for finding the optimal fusion architecture for diverse weather situations. A new training strategy is also introduced such that the performance of the object detector is greatly enhanced in adverse weather scenarios or if a sensor fails. 
	Furthermore, the paper responds to the question if the detection accuracy can be increased further by providing the neural network with a-priori knowledge such as the spatial calibration of the sensors.  
\end{abstract}

\section{Introduction}

\begin{figure}[tbp]
		\centerline{\includegraphics[width=1.0\columnwidth]{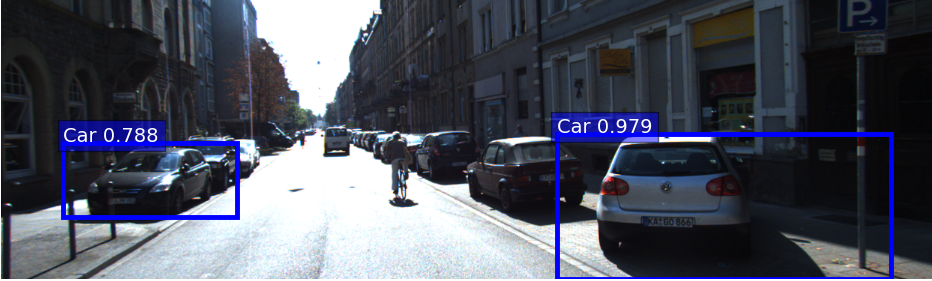}}
		\caption{Result of a state-of-the-art object detector (Faster-RCNN \cite{Ren_2015_Faster_RCNN}) if the lidar sensor fails.}
		\label{fig_Motivation_normal}
	\end{figure}

\begin{figure}[tbp]
		\centerline{\includegraphics[width=1.0\columnwidth]{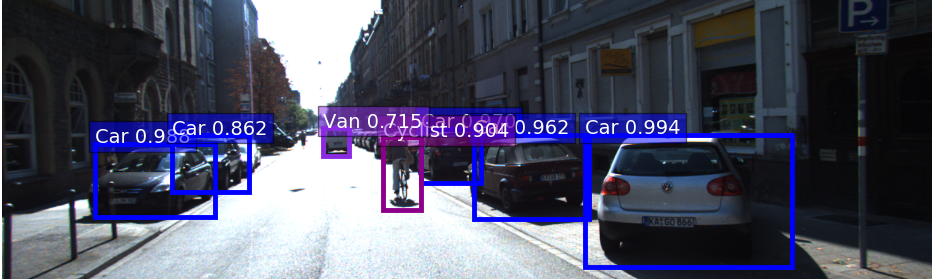}}
		\caption{Result of the proposed object detector if the lidar sensor fails.}
		\label{fig_Motivation_adverse}
	\end{figure}

	A robust object detection is an important basis for intelligent systems, such as domestic robots and autonomous cars. State-of-the-art object detection approaches rely on artificial neural networks, and not only exhibit high performance but are also real-time capable, and applicable in various environments and on a multitude of existing platforms.
	The drawback of object detectors is that they yield good results only in good weather conditions, but their performance decreases enormously in adverse weather conditions, such as snow, rain, fog or even in dazzling sun. 
	Hence, many object detectors make use of several different sensor types to compensate for the weather sensitivity of individual sensors. For instance, in blinding sunlight, camera images contains large white areas without any information about the environment. The lidar sensor still delivers the depth values of the corresponding environment so that obstacles can be detected only by using the lidar data. However, most of the current object detectors \cite{Chen_2016_MultiView3DObjectDetectionNetworkForAutonomousDriving, Ren_2015_Faster_RCNN, Mousavian_2016_3DBoundingBoxEstiamtionUsingDeepLearningAndGeometry} are trained and optimized for good weather conditions greatly depending on good and error-free sensor data.
	In this work, an approach is introduced which is able to deal with partial and complete sensor failure, still delivering good performance under this conditions. The difference between current state-of-the-art and the proposed object detectors can be seen in Fig. \ref{fig_Motivation_normal} and Fig. \ref{fig_Motivation_adverse}, in case the lidar sensor fails. Fig. \ref{fig_Motivation_normal} shows the result of the object detector Faster-RCNN \cite{Ren_2015_Faster_RCNN}, while the result of the proposed detector is illustrated in Fig \ref{fig_Motivation_adverse}.

	A crucial part of the current object detectors is an appropriate sensor data fusion, since many object detection algorithms are based on not only camera images alone, but also consider depth information extracted from lidar or stereo cameras. Generally, there are different possibilities to fuse this different sensor data by means of neural networks. On the one hand, camera and depth image can be fused at the beginning of the object detection approach, such that the network can learn how to merge the different sensors best. On the other hand, each sensor has different characteristics so that it makes sense to determine the features for each sensor separately, and then to combine these sensor-specific features for a further processing.  
	There are various examples in the literature for either fusing the sensor data at the beginning (Early Fusion) \cite{Premebida_2014_PedestrianDetectionCombiningRGBandDenseLidarData, Gonzalez_2017_OnBoardObjectDetectionMulticueMultimodalAndMultiviewRandomForestOfLocalExperts, Gonzalez_2015_MultiviewRandomForestOfLcoalExpertsCombiningRGBandLIDARdataForPedestrianDetection}, or at the end of the fusion process (Late Fusion) \cite{Chen_2016_MultiView3DObjectDetectionNetworkForAutonomousDriving, Mees_2016_ChoosingSmartly_AdaptiveMultimodalFusionForObjectDetectionInChangingEnvironment} so that it is difficult to decide, which sensor fusion approach is suited, especially since the object detectors partially use distinct features.
	Hence, a fair comparison is impossible without an identical test setup. 
	In this work, the different fusion approaches are compared qualitatively by an object detector based on the Faster RCNN approach \cite{Ren_2015_Faster_RCNN} using identical training conditions and the same input features. Furthermore, it is analyzed, whether the optimal fusion architecture for good weather conditions is suitable for adverse weather conditions, too. Different feature representations are tested to answer the question if the performance of the object detector can be increased by equipping the fusion system with a priori knowledge such as the spatial calibration of the sensors.

\section{Related Work}
\label{chap_RelatedWork}

	In recent years, most of the state-of-the-art object detectors use neural networks for an accurate and fast object detection. There are many different object detectors which uses various features and different senor data. However, most of these detectors are a variation or modification of one of the three meta-architectures, namely the Faster-RCNN \cite{Ren_2015_Faster_RCNN}, R-FCN (Region-based Fully Convolutional Networks \cite{Dai_2016_RFCN_ObjectDetectionViaRegionBasedFullyConvolutionalNetworks}) and the SSD (Single Shot Multibox Detector \cite{Liu_2016_SSD_SingleShotMultiBoxDetector}). In the following section, the idea of these meta-architectures is briefly explained, and a short summery about object detectors using various sensor data is given.
	
	The idea of Faster-RCNN \cite{Ren_2015_Faster_RCNN} is to remodel
	the object detection problem as an image classification problem by first determining regions from the image which might contain an object (denoted as Regions of Interest (RoI) or proposals), and by then classifying the object class of the corresponding region. 
	For this, appropriate features are extracted by means of a feature encoder (e.g. VGG16 \cite{Simonyan_2015_VeryDeepConvolutionalNetworksForLargeScaleImageRecognition}) in a first stage, yielding a feature map for further processing. In the second stage, the Region Proposal Network (RPN) introduced in \cite{Ren_2015_Faster_RCNN}  uses this feature map to determine several RoIs (usually 300). For each RoI, a Multilayer Perceptron (MLP) is applied to predict the object class and the offset between proposed and actual bounding box.
	
	A similar architecture to Faster-RCNN is the R-FCN \cite{Dai_2016_RFCN_ObjectDetectionViaRegionBasedFullyConvolutionalNetworks}. However, in contrast to Faster-RCNN, where the MLP must be applied for each RoI separately causing high computational costs, the computations of the second stage are shared on the entire feature map such that the computation time can be reduced. 
		
	Another common meta-architecture for object detection is the SSD \cite{Liu_2016_SSD_SingleShotMultiBoxDetector}. The SSD approach differs from Faster-RCNN and R-FCN in the number of stages, since SSD skips the RPN step and predicts the bounding box and the corresponding object classes in one step. Hence, the inference time is lower than the one for the other two approaches. More details about the pros and cons of these three meta-architectures were analyzed by Huang et al. \cite{Huang_2016_SpeedAccuracyTradeOffsForModernConvolutionalObjectDetectors}, who compared them in relation to accuracy, speed and memory usage. Based on this comparison, the Faster-RCNN was chosen in this work since it outperforms the other meta-architectures with regard to accuracy, while speed and memory usage is not given too much importance. Moreover, it is assumed that they behave in a similar manner in relation to the different proposed sensor fusion architectures.

	Fusing various sensor data for object detection, e.g. mono and stereo camera, or mono-camera and lidar, becomes more and more popular, since the detection rate can be increased further. For instance, Enzweiler et al. \cite{Enzweiler_2011_AMultilevelMixtureOfExpertsFrameworkForPedestrainClassification} detect pedestrians using a mixture-of-experts framework, in which RGB and depth images are fused in the early stage. Chen et al. \cite{chen_2015_3DObjectProposalsForAccurateObjectClassDetection} uses the depth information delivered by a stereo camera to improve the proposal generation and, therefore, the accuracy of the 3D object detection. Mess et al. \cite{Mees_2016_ChoosingSmartly_AdaptiveMultimodalFusionForObjectDetectionInChangingEnvironment} proposed an object detector which smartly chooses the sensor most appropriate for the respective environment.
	There are also approaches using camera images and lidar data. For example, Premebida et al. \cite{Premebida_2014_PedestrianDetectionCombiningRGBandDenseLidarData, Gonzalez_2017_OnBoardObjectDetectionMulticueMultimodalAndMultiviewRandomForestOfLocalExperts, Gonzalez_2015_MultiviewRandomForestOfLcoalExpertsCombiningRGBandLIDARdataForPedestrianDetection} project the lidar points into the image plane coordinates and fills the empty pixels using a bilateral filtering approach. Then, they detect various objects such as pedestrians and cars using random forests methods.
	A popular fusion architecture for object detection is the Multi-View 3D networks (MV3D \cite{Chen_2016_MultiView3DObjectDetectionNetworkForAutonomousDriving}). MV3D combines camera and lidar data for 3D object detection by extracting different features from multiple views, such as birdview and frontview, using intensity, height and density of the lidar points. Each feature is processed separately before they are combined in the end and objects are detected.

\section{Senor Data Preprocessing}

	\begin{figure*}[tb]
		\includegraphics[width=0.95\textwidth]{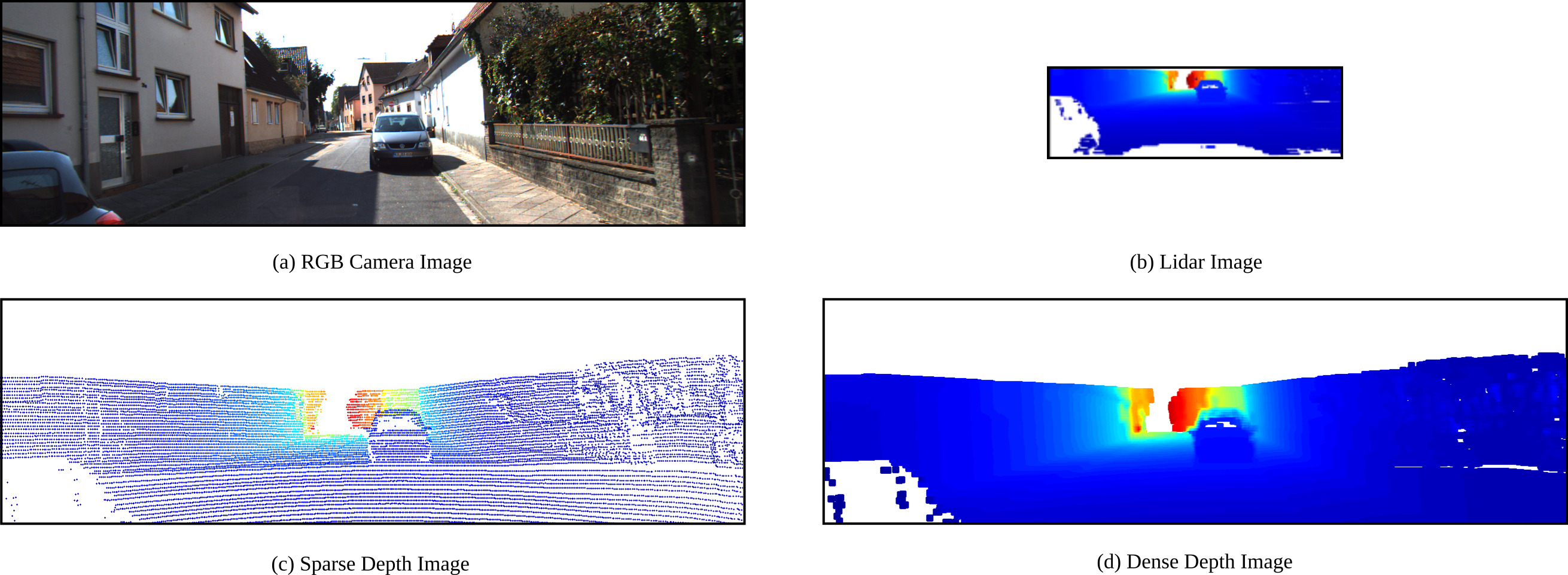}
		\caption{Possible input features for the proposed object detector}
		\label{fig_LidarRepresentations}
	\end{figure*}

	For the fusion of camera and lidar data by means of neural networks, the sensor data have to transformed into a suitable format such that the data can be fed into the neural net. Therefore, some preprocessing steps for camera images and lidar data are necessary. Camera images are usually first resized to a predefined input size of the neural network while keeping its aspect ratio constant. Furthermore, the image mean is subtracted from the RGB image to reduce the influence of the image illumination.	
	For lidar data, the preprocessing steps are more complicated. Usually, the lidar sensor delivers a point cloud $P$ containing $N$ points $\mathbf{p}_i = (x,y,z)^T; i \in 1 \dots N$. In a first step, all points outside the field of view of the camera are removed, since only the data is to be fused in the field of view of the camera in this work. A common problem is that processing 3D point clouds by means of neural networks is time-intensive due to the sparse nature of the 3D points and the high computation costs of 3D convolutions. 
	Hence, most of  the current approaches convert the 3D points into a 2D feature map enabling a fast and good processing by means of neural networks (e.g. \cite{Li_2016_VehicleDetectionfrom3DLidarUsingFullyConvolutionalNetwork, Chen_2016_MultiView3DObjectDetectionNetworkForAutonomousDriving, Gonzalez_2017_OnBoardObjectDetectionMulticueMultimodalAndMultiviewRandomForestOfLocalExperts}). In this work, different 2D lidar representation are considered and evaluated later on. One possibility is to convert the 3D lidar points to a 2D range scan image, denoted as lidar image hereinafter. The lidar image contains no a priori knowledge, e.g. the spatial calibration of the different sensors. However, adding a priori knowledge to the system might improve the performance, since this knowledge does not have to be learned by the neural network. 
	Hence, another possible lidar representation is to project the 3D lidar points into the image plane resulting in a sparse depth map. By this projection, the known spatial relationship of camera and lidar is given to the system. The sparse depth image can be enhanced further by adding additional a priori knowledge yielding a so called dense depth image. The dense depth map is created by interpolating between the sparse lidar points within a small neighborhood. In the following sections, more details about each 2D lidar representation are given.

\subsection{Lidar Image}

	The lidar image only consists of information delivered by the lidar sensor. Its height corresponds to the amount of scanning channels $C$ and its width to the number of points of one scanning rotation. Moreover, each pixel of the lidar image represents depth information. Assuming the lidar sensor delivers a point cloud $P$ of $N$ points $\mathbf{p}_i$, the corresponding lidar image is determined following the approach of Li \cite{Li_2016_VehicleDetectionfrom3DLidarUsingFullyConvolutionalNetwork}. Each point $\mathbf{p}_i$ is transformed by 
	\begin{align}
		%\phi &= arcsin(z / \sqrt{x^2 + y^2 + z^2}) \\
		%\Theta &= atan2(y,x) \\
		%c &= \lfloor \phi / \Delta \phi \rfloor \\
		%r &= \lfloor \Theta / \Delta \Theta \rfloor 
		c &= \lfloor arcsin(z / \sqrt{x^2 + y^2 + z^2}) / \Delta \phi \rfloor \\
		r &= \lfloor atan2(y,x)  / \Delta \Theta \rfloor 
	\end{align}
	to the position $(c,r)$ of the lidar image, where %$\phi$ and $\Theta$ are the elevation and azimuth angle of the 3D point, and 
	$\Delta \phi$ and $\Delta \Theta$ are the average vertical and horizontal angle resolution of the lidar sensor. Each pixel $(c,r)$ of the lidar image is assigned the corresponding depth value $d = \sqrt{x^2 + y^2}$ if there exists a point projected into the corresponding position. Otherwise, no information is available, and hence, the depth value $d$ is set to infinity. A typical example of the lidar image is shown in Fig. \ref{fig_LidarRepresentations}(b).

\subsection{Sparse Depth Image}

	The disadvantage of the lidar image is that camera and lidar images have different sizes. Therefore, it is not possible to merge camera and lidar data to a common 4D input-tensor, as required by the Early Fusion approach.
	In contrast, a suitable 2D representation of the lidar data is a sparse lidar image, which has the same size as the camera image such that camera and lidar data can be combined into a common 4D tensor. For the sparse lidar image, the 3D points are projected to the image plane. By this projection, the neural net is provided with some additional information, namely the spatial relation of camera and lidar sensor by means of the projection matrix. According to \cite{Zhang_2011_CameraCalibarationWithLensDistortionFromLowRankTextures}, the sparse depth image is yielded by
	\begin{align}
		x_n &= - x / z \\
		y_n &= - y / z
	\end{align}
	\begin{align}
		%x_n &= - \frac{x}{z}\\
		%y_n &= - \frac{y}{z} \\
		%x_n &= - x / z\\
		%y_n &= - y / z \\
		r^2 &= x_n^2 + y_n^2 \\
		f(r) &= 1 + \kappa_1 r^2 + \kappa_2 r^4 + \kappa_3 r^6\\
		\mathbf{m}_d &= \left( \begin{array}{c}
						f(r) x_n + 2 \kappa_4 x_n y_n + \kappa_5 \left(r^2 + 2 x_n^2 \right) \\
						f(r) y_n + 2 \kappa_5 x_n y_n + \kappa_4 \left(r^2 + 2 y_n^2 \right) \\
					        1\\
					       \end{array} \right)\\
		\tilde{\mathbf{p}}_{i} &= \left( \begin{array}{ccc}
		                             f_x & \theta & o_x \\
					     0 & f_y & o_y \\
					     0 & 0 & 1 \\
		                            \end{array} \right) \cdot \mathbf{m}_d
	\end{align}
	using a pinhole camera model with lens distortion. Here, $f_x$ and $f_y$ are the focal length along $x$ and $y$-axes, $\left(o_x, o_y \right)$ is the optical center of the camera and $\theta$ its skew parameter. The distortion parameters are denoted as $\kappa_i, i \in \{1,\dots,5\}$. 
	Analogously to the lidar image, each projected point $\tilde{\mathbf{p}}_{i}$ is assigned the depth value $d =  \sqrt{x^2 + y^2}$. All other pixels are set to infinity. An example of this sparse depth image is illustrated in Fig \ref{fig_LidarRepresentations}(c).

\subsection{Dense Depth Image}

	Transforming the 3D lidar points into the image plane leads to a sparse representation, where most of the pixels do not contain any depth information. This might be a problem especially for smaller network architectures, and the performance of the fusion approach might become worse. In the optimal case, each pixel of this representation contains the corresponding depth value similar to a disparity map of stereo cameras. Therefore, further information is added to the system by assuming that the depth values between the sparse projected lidar points do not deviate much. By this, a dense depth map can be generated without any loss of information and the pixels without depth information are filled by means of interpolation. For instance, an easy interpolation method is to take the average depth value of the lidar points within a small neighborhood $\mathcal{N}$, e.g. a square area centered around $\mathbf{p}$:
	\begin{align}
		d^* = \sum_{\mathbf{p}\in \mathcal{N}} \frac{1}{N_{\mathcal{N}}} d_{\mathbf{p}},
	\end{align}
	where $N_{\mathcal{N}}$ is the amount of points within $\mathcal{N}$ and $d^*$ the interpolated depth value. The advantage of this simple interpolation method is its low computation time by the usage of integral images which is a critical point for real-time capable object detectors. If no point lies within the neighborhood $\mathcal{N}$, the corresponding pixel is set to infinity analogously to the lidar image and the sparse depth image. Fig. \ref{fig_LidarRepresentations}(d) shows the determined dense depth map.

\section{Network Architectures}

	In this section, different sensor fusion architectures are described, which are based on the camera-based object detector Faster RCNN \cite{Ren_2015_Faster_RCNN}, and evaluated by means of an object detection task. Thereby, the object detector is extended so that the input data delivered by various sensors, e.g. camera and lidar, can be processed by means of the neural networks.
	As mentioned in Section \ref{chap_RelatedWork}, Faster RCNN consists of two stages: 
	In stage one, suitable features are determined by means of a feature encoder and a feature map is yielded.
	%In stage one, a feature map is determined by means of a feature encoder.
	In the second stage, this feature map is then combined with the RPN, which recognizes and localizes the objects in the image. 
	The actual sensor data fusion takes place in the first stage of the object detector. The different sensor data are processed such that a common feature map is constructed, which is necessary for the second stage. As a basic architecture for the feature encoder serves the VGG16 network architecture \cite{Simonyan_2015_VeryDeepConvolutionalNetworksForLargeScaleImageRecognition}, which consists of five convolutional blocks with a total of 16 convolutional layers.
	The VGG16 architecture is compared with a reduced version of VGG16, the so called VGG16m architecture, which needs less memory and computation time due to the reduced number of convolutional layers. More details about the net architectures of VGG16 and VGG16m can be found in Table \ref{tabel_networkArchitecture}.
	In this work, different fusion architectures are examined in order to find the best fusion strategy. For instance, the senor data can be fused right at the beginning, or at the end of the feature encoder. These architectures are called Early Fusion and Late Fusion respectively, and are investigated more closely in the next sections.

	\begin{table}[tb]
		\caption{basic network architecture of the used feature encoders: The convN-C denotes a convolutional layer with kernel size N and C output channels. For simplicity, the ReLu activation function are neglected here.}
		\begin{center}
			\begin{tabular}{|c|c|c|}
				\hline
				\textbf{conv. block} & \textbf{VGG16m}& \textbf{VGG16} \Tstrut \Bstrut \\ \hline \Tstrut
				1 & conv7-96 - stride 2 & conv3-64 - stride 1 \Mstrut \\
				& maxpooling - stride 2 & conv3-64 - stride 1 \Mstrut \\
				& & maxpooling - stride 2 \Bstrut \\ \hline \Tstrut
				2 & conv5-256 - stride 2 & conv3-128 - stride 1 \Mstrut \\
				& maxpooling - stride 2 & conv3-128 - stride 1 \Mstrut \\ 
				& & maxpooling - stride 2 \Bstrut \\ \hline \Tstrut
				3 & conv3-516 - stride 1 & conv3-256 - stride 1 \Mstrut \\
				& & conv3-256 - stride 1 \Mstrut \\ 
				& & conv3-256 - stride 1 \Mstrut \\
				& & maxpooling - stride 2 \Bstrut \\ \hline \Tstrut
				4 & conv3-516 - stride 1 & conv3-516 - stride 1 \Mstrut \\
				& & conv3-516 - stride 1 \Mstrut \\ 
				& & conv3-516 - stride 1 \Mstrut \\
				& & maxpooling - stride 2 \Bstrut \\ \hline \Tstrut
				5 & conv3-516 - stride 1 & conv3-516 - stride 1 \Mstrut \\
				& & conv3-516 - stride 1 \Mstrut \\ 
				& & conv3-516 - stride 1 \Bstrut \\ \hline
			\end{tabular}
			\label{tabel_networkArchitecture}
		\end{center}
	\end{table}

\subsection{Early Fusion}

\begin{figure*}[tbp]
		\includegraphics[width=1.0\textwidth]{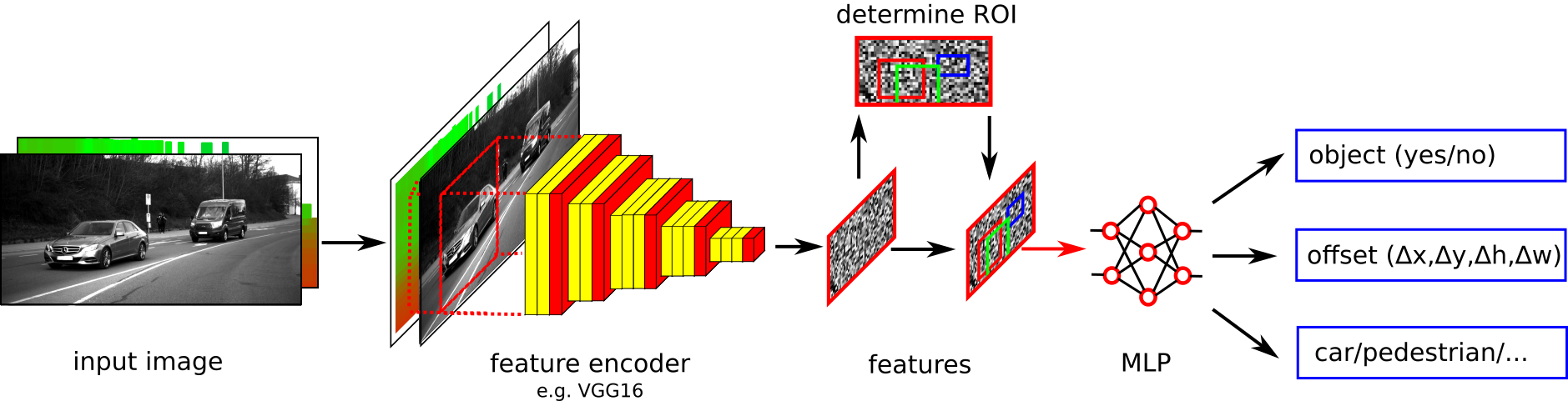}
		\caption{Early Fusion Approach: A common feature map is determined for all sensors by the feature encoder. For this, RGB image and the depth image determined from the lidar data are fit into a 4D tensor and progressed together in the neural network}
		\label{fig_EarlyFusion}
	\end{figure*}

	The idea of the Early Fusion approach is that the neural network should learn how to fuse the different sensor data without any human engineered features or special net configurations. Instead, the sensor data are combined to a common input tensor and processed together in the entire network. Therefore, the input dimension of the camera-based object detector is extended from a 3D tensor to a 4D tensor, where the first three dimensions contain the three color channels of the RGB image, and the fourth dimension is the preprocessed lidar data. 
	A feature encoder is applied on the created input tensor, and a common feature map is yielded, which is required for the second stage of the object detector. 
	The described net architecture is shown in Fig \ref{fig_EarlyFusion}. Note, that only the sparse and dense depth image are suitable for the Early Fusion approach, since the size of the lidar image is much smaller than the size of the camera image, and hence, they cannot be fitted into a common tensor.

	\begin{figure*}[tbp]
		\includegraphics[width=1.0\textwidth]{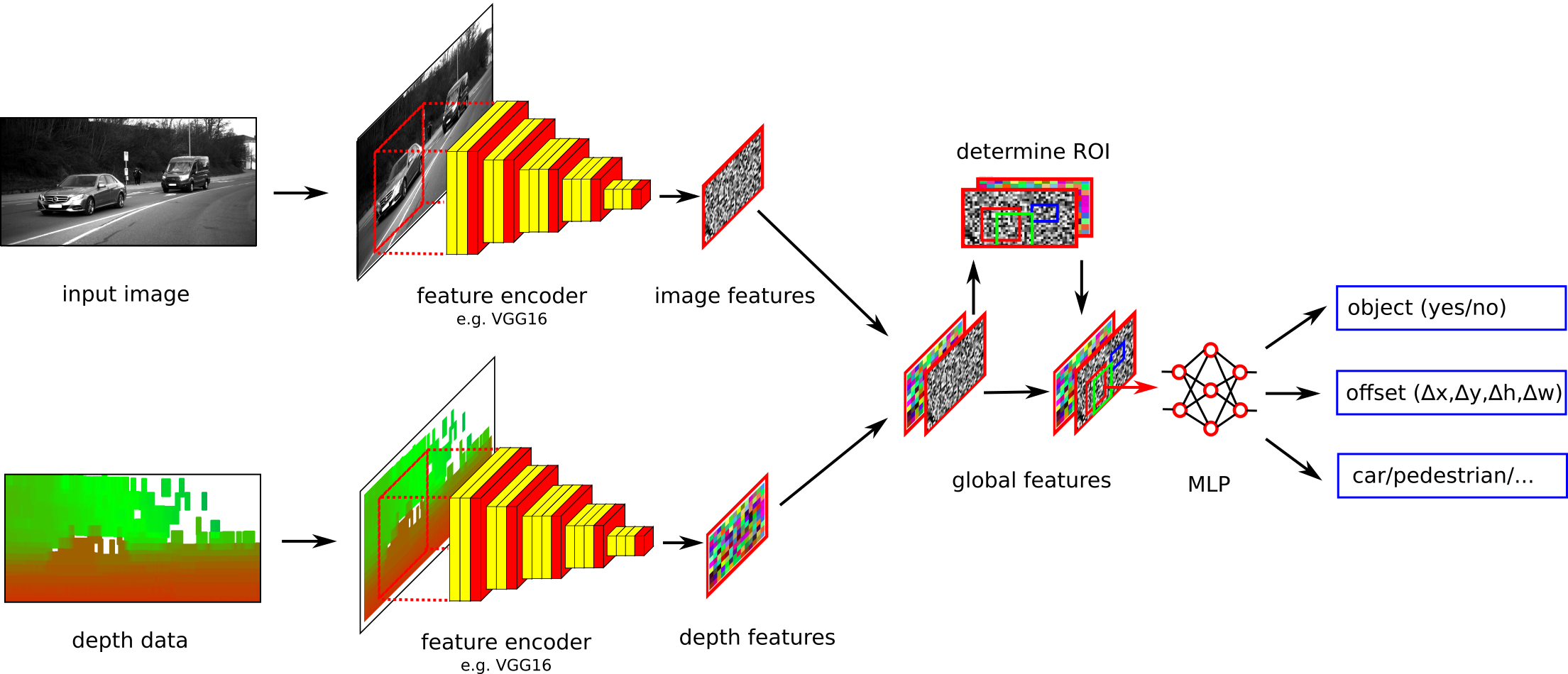}
		\caption{Late Fusion Approach: For each sensor, an independent feature map is determined by distinct feature encoder. The camera and lidar feature maps are then concatenated before the RPN is applied.}
		\label{fig_LateFusion}
	\end{figure*}

\subsection{Late Fusion}

	The disadvantage of Early Fusion is that the outputs of the different sensor types vary, and each sensor has sensor specific properties. These sensor specific properties might be lost by an early fusion. Hence, another net architecture is considered which overcomes this problem by first processing the sensor data separately before they are finally fused. More precisely, for each sensor a separate feature map is determined by means of different feature encoders. These feature maps are then combined to a common feature map, which is the input of the second stage of the object detector. The Late Fusion approach, whose net architecture is illustrated in Fig. \ref{fig_LateFusion}, also allows the use of different dimensions of input data so that the lidar image can be used for this architecture, too.

\subsection{Middle Fusion}

	The Early and Late Fusion approaches exhibit different advantages and disadvantages. Therefore, it makes sense to design yet another architecture, which is a combination of both, the so called Middle Fusion. By this approach, the distinct sensor data are first processed independently from each other, and are concatenated at a later stage. Then, the concatenated sensor data are processed further by the neural network yielding a common feature map which is the starting point for the RPN. 	
	In this particular case, camera and lidar data are treated separately up to convolutional block 3, and starting from convolutional block 4, all sensor data are processed together.
	The Middle Fusion approach enables the network to learn sensor specific properties and then how to combine these properties to suitable features for a good and robust object detection.

\section{Evaluation}

	\begin{table*}[tb]
		\caption{Evaluation at Good Weather Conditions (VGG16m)}
		\begin{center}
			\begin{tabular}{|c|c||c|c|c|c|c|c|c||c|}
				\hline
				\multicolumn{2}{|c||}{\textbf{approach}} &  \multicolumn{7}{|c||}{\textbf{classes}} & \textbf{computation time} \Tstrut \Bstrut \\ \cline{1-9}
				fusion approach & input data & car & truck & tram & pedestrian & cyclist & van & \textbf{mAP} & \Tstrut \Bstrut \\ \hline \hline \Tstrut
				none & RGB & $0.741$ & $0.817$ & $0.678$ & $0.417$ & $0.587$ & $0.672$ & $\mathbf{0.651}$ & $36.64$ms  \Bstrut \\ \hline \Tstrut
				%& sparse depth image & $0.717$ & $0.671$ & $0.544$ & $0.364$ & $0.370$ & $0.491$ & $\mathbf{0.526}$ & $27.80$ms \\ 
				%& dense depth image &  $0.717$ & $0.710$ & $0.549$ & $0.411$ & $0.498$ & $0.526$ & $\mathbf{0.568}$ & $27.55$ms \\ \hline
				Early Fusion & RGB \& lidar image$^{\mathrm{*}}$ & $-$ & $-$ & $-$ & $-$ & $-$ & $-$ & $-$ & $-$ \Mstrut \\
				& RGB \& sparse depth image & $0.751$ & $0.842$ & $0.725$ & $0.433$ & $0.593$ & $0.690$ & $\mathbf{0.672}$ & $43.20$ms \Mstrut \\ 
				& RGB \& dense depth image & $0.746$ & $0.850$ & $0.740$ & $0.426$ & $0.566$ & $0.702$ & $\mathbf{0.673}$ & $43.15$ms  \Bstrut\\ \hline \Tstrut
				Middle Fusion & RGB \& lidar image & $0.740$ & $0.735$ & $0.641$ & $0.419$ & $0.579$ & $0.642$ & $\mathbf{0.626}$  & $43.47$ms \Mstrut \\
				& RGB \& sparse depth image & $0.750$ & $0.822$ & $0.733$ & $0.479$ & $0.595$ & $0.665$ & $\mathbf{0.674}$ & $45.69$ms \Mstrut \\ 
				& RGB \& dense depth image & $0.751$ & $0.831$ & $0.751$ & $0.442$ & $0.611$ & $0.697$ & $\mathbf{0.680}$ & $46.49$ms  \Bstrut\\ \hline \Tstrut
				Late Fusion & RGB \& lidar image & $0.745$ & $0.765$ & $0.736$ & $0.419$ & $0.570$ & $0.638$ & $\mathbf{0.646}$ & $49.57$ms \Mstrut \\
				& RGB \& sparse depth image & $0.745$ &  $0.829$ & $0.737$ & $0.445$ & $0.548$ & $0.658$ & $\mathbf{0.660}$ & $52.53$ms \Mstrut \\ 
				& RGB \& dense depth image & $0.746$ & $0.836$ & $0.760$ & $0.432$ & $0.565$ & $0.711$ & $\mathbf{0.675}$ & $52.75$ms \Bstrut\\ \hline   
				\multicolumn{10}{l}{$^{\mathrm{*}}$The lidar image is not suitable for the Early Fusion approach, since its size is different to the image size.}\Tstrut 
			\end{tabular}
			\label{tabel_resultVGG16m_good}
		\end{center}
	\end{table*}
	
	\begin{table*}
		\caption{Evaluation at Good Weather Conditions (VGG16)}
		\begin{center}
			\begin{tabular}{|c|c||c|c|c|c|c|c|c||c|}
				\hline
				\multicolumn{2}{|c||}{\textbf{approach}} &  \multicolumn{7}{|c||}{\textbf{classes}} & \textbf{computation time} \Tstrut \Bstrut \\ \cline{1-9}
				fusion approach & input data & car & truck & tram & pedestrian & cyclist & van & \textbf{mAP} & \Tstrut \Bstrut \\ \hline \hline \Tstrut
				none & RGB & $0.782$ & $0.883$ & $0.878$ & $0.505$ & $0.693$ & $0.780$ & $\mathbf{0.753}$  & $64.54$ms \Bstrut \\ \hline \Tstrut
				%& sparse depth image & $0.$ & $0.$ & $0.$ & $0.$ & $0.$ & $0.$ & $\mathbf{0.}$ & $57.80$ms \\ 
				%& dense depth image & $0.$ & $0.$ & $0.$ & $0.$ & $0.$ & $0.$ & $\mathbf{0.}$ & $57.53$ms \\ \hline
				Early Fusion & RGB \& lidar image$^{\mathrm{*}}$ & $-$ & $-$ & $-$ & $-$ & $-$ & $-$ & $-$ & $-$ \Mstrut \\
				& RGB \& sparse depth image & $0.784$ & $0.888$ & $0.800$ & $0.558$ & $0.709$ & $0.785$ & $\mathbf{0.757}$  & $70.98$ms \Mstrut \\ 
				& RGB \& dense depth image & $0.776$ & $0.912$ & $0.871$ & $0.509$ & $0.729$ & $0.778$ & $\mathbf{0.762}$ & $70.98$ms \Bstrut\\ \hline \Tstrut
				Middle Fusion & RGB \& lidar image$^{\mathrm{**}}$ & $-$ & $-$ & $-$ & $-$ & $-$ & $-$ & $-$ & $-$  \Mstrut \\
				& RGB \& sparse depth image & $0.781$ & $0.899$ & $0.872$ & $0.517$ & $0.699$ & $0.767$ & $\mathbf{0.756}$ & $94.57$ms \Mstrut \\ 
				& RGB \& dense depth image & $0.780$ & $0.908$ & $0.849$ & $0.511$ & $0.714$ & $0.774$ & $\mathbf{0.756}$ & $94.69$ms \Bstrut\\ \hline \Tstrut
				Late Fusion & RGB \& lidar image & $0.780$ & $0.899$ & $0.881$ & $0.514$ & $0.687$ & $0.778$ & $\mathbf{0.756}$ & $90.51$ms \Mstrut \\
				& RGB \& sparse depth image & $0.776$ & $0.876$ & $0.864$ & $0.512$ & $0.708$ & $0.766$ & $\mathbf{0.754}$ & $115.95$ms \Mstrut \\ 
				& RGB \& dense depth image & $0.777$ & $0.885$ & $0.861$ & $0.516$ & $0.712$ & $0.782$ & $\mathbf{0.758}$ & $115.43$ms \Bstrut\\ \hline 
				\multicolumn{10}{l}{$^{\mathrm{*}}$The lidar image is not suitable for the Early Fusion approach, since its size is different to the image size.} \Tstrut \\
				\multicolumn{10}{l}{$^{\mathrm{**}}$The lidar image is not suitable for the Middle Fusion approach, since the feature maps of conv3 cannot be concatenated due to different output size}\\
			\end{tabular}
			\label{tabel_resultVGG16_good}
		\end{center}
	\end{table*}

	In the previous sections, different network architectures for sensor data fusion and various input representations of the lidar data are described. Now, these different approaches are compared and evaluated qualitatively on the Kitti object detection benchmark \cite{kitti_Dataset} in terms of accuracy and computation time. The Kitti dataset consists of 7481 training images and 7518 test images with the corresponding lidar data recorded by a Velodyne 64. In total, the annotated dataset contains about 80k labeled objects in seven different classes (car, truck, tram, pedestrian, cyclist, van). In contrast to many other works, e.g. \cite{Chen_2016_MultiView3DObjectDetectionNetworkForAutonomousDriving, Yang_2016_ExploitAlltheLayersFastAndAccurateCNNObjectDetectorwithScaleDependentPoolingandCascadedRejectionClassifiers, Mousavian_2016_3DBoundingBoxEstiamtionUsingDeepLearningAndGeometry}, where the evaluation is focused on detecting cars, pedestrians and cyclists, the detection accuracy of the proposed approaches is determined for all seven categories. 
	The benchmark offers three levels of difficulty, namely easy, moderate and hard, which differs in occlusion and truncation, however, only the most challenging part is considered in this work. Since the %annotated 
	ground-truth objects of the test set are not publicly available, the training set is split into a training, validation and a test set. The test set is composed of the first 500 images of the training set, the training set of the next 6500 images and the validation set of the remaining images. 
	
	In this work, the focus is set on identical training conditions for all proposed approaches so that only their different architectures are compared. Each approach is implemented in Caffe \cite{caffe} and trained on the training set using a batch size of $4$ and a momentum of $0.9$. The training loss is minimized by Stochastic Gradient Descent (SGD), and the initial learning rate is set to $0.001$. After every $50k$ training iterations, the learning rate is reduced by a factor of $10$. The validation set is used to monitor the training process for avoiding overfitting and to abort the training process if the training and validation loss differ too much. Usually, the training process is aborted after $100k$ to $150k$ iterations. Each training process is executed three times and the average of the results is taken in order to compensate training fluctuations. 
	The training parameters are initialized by a pretrained VGG16 net on the ImageNet dataset. Since the input dimensions of Early, Middle and Late Fusion are different, the parameters of the first convolutional layer are initialized randomly using a Gaussian distribution. Finally, the different approaches are compared on the test set using the average precision (AP) as an evaluation metric. %The average precision is defined as the ratio of the number of the detected objects, which are detected correctly according to the ground-truth. 
	The evaluation results in good and adverse weather conditions can be found in the next sections.

\subsection{Evaluation at Good Weather Conditions}

	\begin{figure*}[tbp]
		\includegraphics[width=1.0\textwidth]{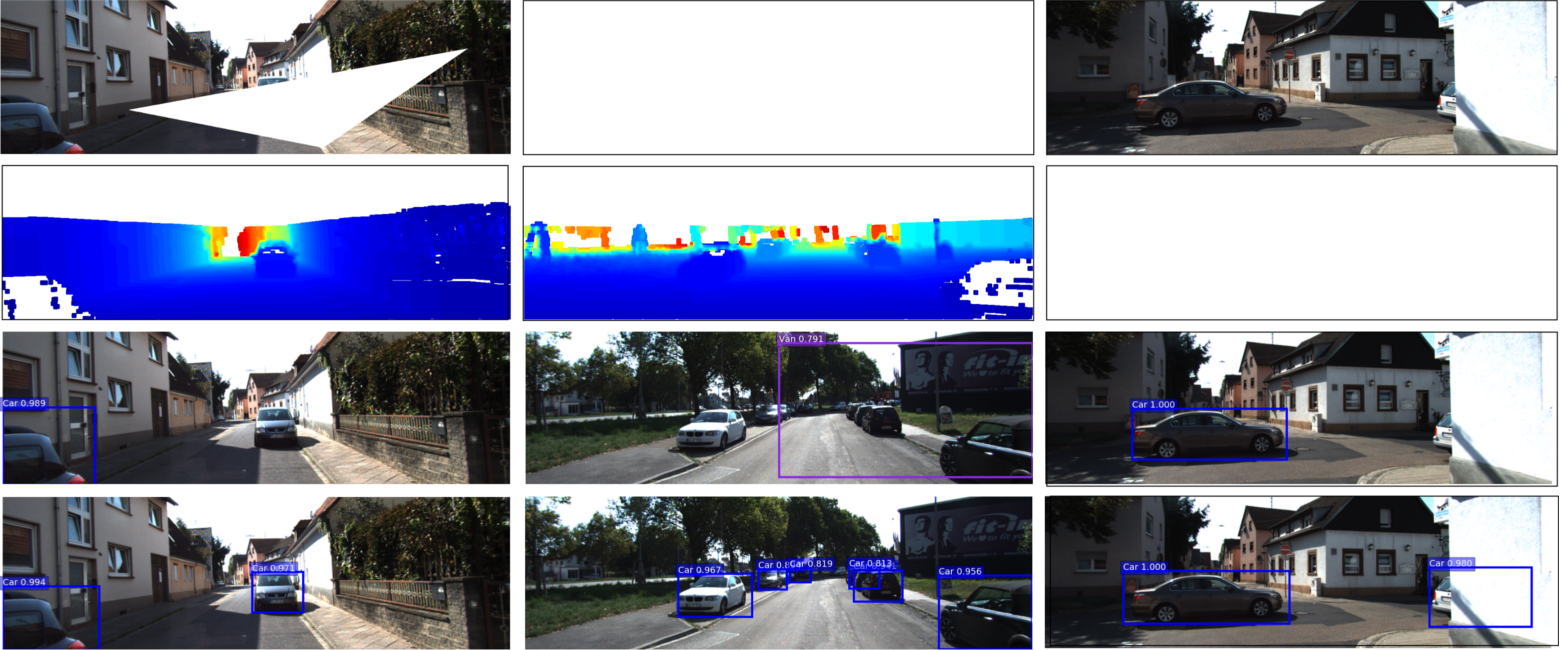}
		\caption{Qualitative results on the Kitti dataset. First line: input image; second line: input dense depth image; third line: proposed object detector trained on Kitti; fourth line: proposed object detector trained on adverse-Kitti (Late Fusion, VGG16).}
		\label{fig_Result}
	\end{figure*}

	 The various, proposed approaches are now evaluated in good weather conditions by means of detection accuracy and computation time, and are compared with the classical object detector using only camera images. %Furthermore, the different input features of the lidar data are compared. 
	 Table \ref{tabel_resultVGG16m_good} and Table \ref{tabel_resultVGG16_good} contain the results of all considered approaches using VGG16m and VGG16 respectively as basis architecture. The results indicate that object detectors benefit from a sensor data fusion, since most of the proposed object detectors outperform the classical object detectors processing images only. As expected, the object detectors based on VGG16 have a greater detection rate than the ones based on VGG16m, because VGG16 consists of more convolutional layers so that the network is able to learn the object attributes better. 
	 Furthermore, Early, Middle and Late Fusion perform similarly good in optimal weather conditions. Their accuracy differs only by a maximum of $0.7\%$ on the Kitti dataset.
	 
	 When comparing the different input data of the lidar data, differences become visible. In case of VGG16m, the dense depth image outperforms the other two lidar representations for all fusion approaches. In contrast, they yield similar results in case of VGG16. The reason is that the smaller feature encoder VGG16m benefits from the added a-priori knowledge while the VGG16 network architecture is large enough to learn this knowledge itself during the training process. 
	 
	 Another important issue is the inference time of the neural nets including its post-processing steps, e.g. applying the Non Maximum Suppression (NMS). The computation time is also listed in Table \ref{tabel_resultVGG16m_good} and Table \ref{tabel_resultVGG16_good} processing input images of size $302 \times 1000$ on a Nvidia Titan X. Generally, the object detectors based on VGG16m are faster than the ones based on VGG16, since VGG16m consists of less layers so that less arithmetic operations are necessary. Similarly, the inference time of the Late Fusion is larger than the one of Middle and Early Fusion for the same reasons. Moreover, the choice of the lidar input features also effects the computation time due to the different input size of the lidar features. The lidar image has the smallest input size, hence, the approaches using the lidar image are faster than the other ones.

\subsection{Evaluation at Adverse Weather Conditions}

	Before evaluating the proposed approaches in adverse weather conditions, the Kitti dataset is modified, since it only consists of road scenes recorded in good weather conditions and no other appropriate adverse weather dataset is known. Hereinafter, this created dataset is called adverse-Kitti. The adverse-Kitti dataset consists of modified Kitti data  apart from the original Kitti data. For instance, white areas are randomly fitted into camera and depth images. 
	This is motivated by the fact that at dazzling sun, the camera images contain white spots, and the lidar sensors deliver less points in rain or snow due to reflection and absorption of the laser beams. Furthermore, a sensor failure is simulated by using only either camera or lidar data while setting the other input stream to infinity. In total, the ratio between good weather data, partly disturbed data and completely impaired data is $1:2:4$. Some example data of adverse-Kitti are shown in Fig. \ref{fig_Result}.

	First, the neural networks trained in the last section in good adverse weather conditions are evaluated by means of adverse-Kitti. For simplicity, the evaluation is restricted to the usage of dense depth images, since it outperforms lidar image and sparse depth image, as shown in the previous section. The results are presented in the Table \ref{tabel_resultadverseKitti_trainedgood} for both feature encoders. It turns out that the performance of the object detector measured in mAP decreases over $20\%$ in adverse weather conditions compared to good weather conditions, since the network parameters are highly optimized for good weather conditions. In contrast to optimal weather conditions, where the different fusion approaches perform similarly good, the Late Fusion approach delivers the best results in adverse weather conditions. The reason is that Late Fusion consists of independent feature encoders which are not disturbed by the sensor failure of the other sensors.
	
	The performance of the object detector in non-optimal weather conditions can be increased by training the different fusion approaches on adverse-Kitti. Table \ref{tabel_resultadverseKitti} shows the results of the evaluation on Kitti and adverse-Kitti using VGG16m and VGG16 respectively as feature encoder. 
	Basically, VGG16m and VGG16 behave in a similar manner. The accuracy on adverse-Kitti increases about $20\%$, however in the case of VGG16m at the expense of the accuracy at good weather conditions. In particular, the performance decreases about $5\%$ on the Kitti dataset compared to networks only trained with good weather data. In contrast, the performance of VGG16 even increases, since the networks learn to trust the individual sensors better. 
	The results also show that Late Fusion outperforms the other fusion approaches. Generally, it yields that the performance of the object detector in adverse weather conditions increases the later the sensor data are fused. This is caused by the fact that the sensor data influence each other at an early fusion. Hence, if one sensor is disturbed, the complete feature encoder is disturbed, and the accuracy of the object detector decreases. 
	In Fig. \ref{fig_Result}, the difference between the object detector trained on Kitti and the object detector trained on adverse-Kitti is shown by the example of the Late Fusion. %approach. 

	\begin{table*}
		\caption{Evaluation on adverse-Kitti using a net trained with Kitti}
		\begin{center}
			\begin{tabular}{|c|c||c|c|c|c|c|c|c|c|}
				\hline
				\multicolumn{2}{|c||}{\textbf{approach}} &  \multicolumn{7}{|c|}{\textbf{classes}}  \Tstrut \Bstrut \\ \hline
				feature encoder & fusion approach & car & truck & tram & pedestrian & cyclist & van & \textbf{mAP} \Tstrut \Bstrut \\ \hline \hline \Tstrut
				VGG16m & Early Fusion & $0.524$ & $0.451$ & $0.410$ & $0.271$ & $0.333$ & $0.408$ & $\mathbf{0.407}$ \Mstrut \\
				& Middle Fusion & $0.545$ & $0.528$ & $0.418$ & $0.272$ & $0.328$ & $0.440$ & $\mathbf{0.422}$ \Mstrut \\
				& Late Fusion & $0.547$ & $0.538$ & $0.415$ & $0.279$ & $0.370$ & $0.455$ & $\mathbf{0.434}$ \Bstrut \\ \hline \Tstrut
				VGG16 & Early Fusion & $0.610$ & $0.671$ & $0.592$ & $0.381$ & $0.505$ & $0.575$ & $\mathbf{0.556}$ \Mstrut \\
				& Middle Fusion & $0.609$ & $0.668$ & $0.590$ & $0.400$ & $0.516$ & $0.568$ & $\mathbf{0.559}$ \Mstrut \\ 
				& Late Fusion & $0.607$ & $0.673$ & $0.632$ & $0.395$ & $0.538$ & $0.583$ & $\mathbf{0.571}$ \Bstrut \\ \hline
			\end{tabular}
			\label{tabel_resultadverseKitti_trainedgood}
		\end{center}
	\end{table*}

	\begin{table*}
		\caption{Evaluation using a net trained with adverse-Kitti}
		\begin{center}
			\begin{tabular}{|c|c|c||c|c|c|c|c|c|c|c|}
				\hline
				\multicolumn{3}{|c||}{\textbf{approach}} &  \multicolumn{7}{|c|}{\textbf{classes}} \Tstrut \Bstrut \\ \hline
				feature encoder & fusion approach & dataset & car & truck & tram & pedestrian & cyclist & van & \textbf{mAP} \Tstrut \Bstrut \\ \hline \hline \Tstrut
				VGG16m & Early Fusion & Kitti & $0.732$ & $0.784$ & $0.588$ & $0.390$ & $0.535$ & $0.621$ & $\mathbf{0.609}$ \Mstrut \\
				& & adverse-Kitti & $0.689$ & $0.657$ & $0.433$ & $0.343$ & $0.413$ & $0.522$ & $\mathbf{0.510}$ \Bstrut \\ \cline{2-10} \Tstrut
				& Middle Fusion & Kitti & $0.738$ & $0.775$ & $0.606$ & $0.414$ & $0.541$ & $0.628$ & $\mathbf{0.617}$ \Mstrut \\
				& & adverse-Kitti & $0.697$ & $0.667$ & $0.474$ & $0.363$ & $0.431$ & $0.539$ & $\mathbf{0.529}$ \Bstrut \\ \cline{2-10} \Tstrut
				& Late Fusion & Kitti & $0.736$ & $0.765$ & $0.681$ & $0.425$ & $0.519$ & $0.644$ & $\mathbf{0.628}$ \Mstrut \\
				& & adverse-Kitti & $0.698$ & $0.663$ & $0.488$ & $0.372$ & $0.408$ & $0.547$ & $\mathbf{0.529}$ \Bstrut \\ \hline \Tstrut
				VGG16 & Early Fusion & Kitti & $0.795$ & $0.899$ & $0.888$ & $0.538$ & $0.752$ & $0.806$ & $\mathbf{0.780}$ \Mstrut \\
				& & adverse-Kitti & $0.766$ & $0.826$ & $0.745$ & $0.498$ & $0.679$ & $0.719$ & $\mathbf{0.707}$ \Bstrut \\ \cline{2-10} \Tstrut
				& Middle Fusion & Kitti & $0.801$ & $0.922$ & $0.915$ & $0.522$ & $0.794$ & $0.817$ & $\mathbf{0.795}$ \Mstrut \\
				& & adverse-Kitti & $0.772$ & $0.856$ & $0.778$ & $0.495$ & $0.705$ & $0.732$ & $\mathbf{0.723}$ \Bstrut \\ \cline{2-10} \Tstrut
				& Late Fusion & Kitti & $0.798$ & $0.907$ & $0.962$ & $0.556$ & $0.758$ & $0.839$ & $\mathbf{0.805}$ \Mstrut \\
				& & adverse-Kitti & $0.766$ & $0.851$ & $0.801$ & $0.506$ & $0.690$ & $0.758$ & $\mathbf{0.729}$ \Bstrut \\ \hline
			\end{tabular}
			\label{tabel_resultadverseKitti}
		\end{center}
	\end{table*}

\section{Conclusion}

	In this paper, the advantages and disadvantages of different fusion architectures were investigated by the task of object detection in diverse weather conditions. It turned out that the later the sensor data is fused, the greater the detection rate of the object detectors is.
	The performance is enhanced further by adding a-priori knowledge to the neural net, especially for small neural networks. 
	Several approaches were trained on the Kitti dataset, and it was shown that their performance decreases rapidly, if they are applied to adverse weather conditions. The performance in adverse weather conditions could be increased by training the network%'s parameters 
	by means of an adverse weather dataset.  
	In good weather conditions, the proposed approach even outperforms object detectors trained on pure good weather conditions, if its feature encoder is large enough.

\section{Acknowledgment}
	
	The research leading to these results has received funding from the European Union under the H2020 ECSEL Programme as part of the DENSE project, contract number 692449.

\bibliography{/home/andreas/Documents/Literatur/Jabref-Datebase/Literatur_Promotion}

\begin{thebibliography}{10}

\bibitem{chen_2015_3DObjectProposalsForAccurateObjectClassDetection}
Xiaozhi Chen, Kaustav Kundu, Yukun Zhu, Andrew Berneshawi, Huimin Ma, Sanja
  Fidler, and Raquel Urtasun.
\newblock 3d object proposals for accurate object class detection.
\newblock In {\em NIPS}, 2015.

\bibitem{Chen_2016_MultiView3DObjectDetectionNetworkForAutonomousDriving}
Xiaozhi Chen, Huimin Ma, Ji~Wan, Bo~Li, and Tian Xia.
\newblock Multi-view 3d object detection network for autonomous driving.
\newblock {\em CoRR}, abs/1611.07759, 2016.

\bibitem{Dai_2016_RFCN_ObjectDetectionViaRegionBasedFullyConvolutionalNetworks}
Jifeng Dai, Yi~Li, Kaiming He, and Jian Sun.
\newblock {R-FCN:} object detection via region-based fully convolutional
  networks.
\newblock {\em CoRR}, abs/1605.06409, 2016.

\bibitem{Enzweiler_2011_AMultilevelMixtureOfExpertsFrameworkForPedestrainClassification}
M.~Enzweiler and D.~M. Gavrila.
\newblock A multilevel mixture-of-experts framework for pedestrian
  classification.
\newblock {\em IEEE Transactions on Image Processing}, 20(10):2967--2979, Oct
  2011.

\bibitem{kitti_Dataset}
Andreas Geiger, Philip Lenz, and Raquel Urtasun.
\newblock Are we ready for autonomous driving? the kitti vision benchmark
  suite.
\newblock In {\em Conference on Computer Vision and Pattern Recognition
  (CVPR)}, 2012.

\bibitem{Gonzalez_2015_MultiviewRandomForestOfLcoalExpertsCombiningRGBandLIDARdataForPedestrianDetection}
A.~González, G.~Villalonga, J.~Xu, D.~Vázquez, J.~Amores, and A.~M. López.
\newblock Multiview random forest of local experts combining rgb and lidar data
  for pedestrian detection.
\newblock In {\em 2015 IEEE Intelligent Vehicles Symposium (IV)}, pages
  356--361, June 2015.

\bibitem{Gonzalez_2017_OnBoardObjectDetectionMulticueMultimodalAndMultiviewRandomForestOfLocalExperts}
A.~González, D.~Vázquez, A.~M. Lóopez, and J.~Amores.
\newblock On-board object detection: Multicue, multimodal, and multiview random
  forest of local experts.
\newblock {\em IEEE Transactions on Cybernetics}, PP(99):1--11, 2017.

\bibitem{Huang_2016_SpeedAccuracyTradeOffsForModernConvolutionalObjectDetectors}
Jonathan Huang, Vivek Rathod, Chen Sun, Menglong Zhu, Anoop Korattikara,
  Alireza Fathi, Ian Fischer, Zbigniew Wojna, Yang Song, Sergio Guadarrama, and
  Kevin Murphy.
\newblock Speed/accuracy trade-offs for modern convolutional object detectors.
\newblock {\em CoRR}, abs/1611.10012, 2016.

\bibitem{caffe}
Yangqing Jia, Evan Shelhamer, Jeff Donahue, Sergey Karayev, Jonathan Long, Ross
  Girshick, Sergio Guadarrama, and Trevor Darrell.
\newblock Caffe: Convolutional architecture for fast feature embedding.
\newblock {\em arXiv preprint arXiv:1408.5093}, 2014.

\bibitem{Li_2016_VehicleDetectionfrom3DLidarUsingFullyConvolutionalNetwork}
Bo~Li, Tianlei Zhang, and Tian Xia.
\newblock Vehicle detection from 3d lidar using fully convolutional network.
\newblock {\em CoRR}, abs/1608.07916, 2016.

\bibitem{Liu_2016_SSD_SingleShotMultiBoxDetector}
Wei Liu, Dragomir Anguelov, Dumitru Erhan, Christian Szegedy, Scott~E. Reed,
  Cheng{-}Yang Fu, and Alexander~C. Berg.
\newblock {SSD:} single shot multibox detector.
\newblock {\em CoRR}, abs/1512.02325, 2015.

\bibitem{Mees_2016_ChoosingSmartly_AdaptiveMultimodalFusionForObjectDetectionInChangingEnvironment}
O.~Mees, A.~Eitel, and W.~Burgard.
\newblock Choosing smartly: Adaptive multimodal fusion for object detection in
  changing environments.
\newblock In {\em 2016 IEEE/RSJ International Conference on Intelligent Robots
  and Systems (IROS)}, pages 151--156, Oct 2016.

\bibitem{Mousavian_2016_3DBoundingBoxEstiamtionUsingDeepLearningAndGeometry}
Arsalan Mousavian, Dragomir Anguelov, John Flynn, and Jana Kosecka.
\newblock 3d bounding box estimation using deep learning and geometry.
\newblock {\em CoRR}, abs/1612.00496, 2016.

\bibitem{Premebida_2014_PedestrianDetectionCombiningRGBandDenseLidarData}
C.~Premebida, J.~Carreira, J.~Batista, and U.~Nunes.
\newblock Pedestrian detection combining rgb and dense lidar data.
\newblock In {\em 2014 IEEE/RSJ International Conference on Intelligent Robots
  and Systems}, pages 4112--4117, Sept 2014.

\bibitem{Ren_2015_Faster_RCNN}
Shaoqing Ren, Kaiming He, Ross~B. Girshick, and Jian Sun.
\newblock Faster {R-CNN:} towards real-time object detection with region
  proposal networks.
\newblock {\em CoRR}, abs/1506.01497, 2015.

\bibitem{Simonyan_2015_VeryDeepConvolutionalNetworksForLargeScaleImageRecognition}
Karen Simonyan and Andrew Zisserman.
\newblock Very deep convolutional networks for large-scale image recognition.
\newblock {\em CoRR}, abs/1409.1556, 2014.

\bibitem{Yang_2016_ExploitAlltheLayersFastAndAccurateCNNObjectDetectorwithScaleDependentPoolingandCascadedRejectionClassifiers}
F.~Yang, W.~Choi, and Y.~Lin.
\newblock Exploit all the layers: Fast and accurate cnn object detector with
  scale dependent pooling and cascaded rejection classifiers.
\newblock In {\em 2016 IEEE Conference on Computer Vision and Pattern
  Recognition (CVPR)}, pages 2129--2137, June 2016.

\bibitem{Zhang_2011_CameraCalibarationWithLensDistortionFromLowRankTextures}
Zhengdong Zhang, Yasuyuki Matsushita, and Yi~Ma.
\newblock Camera calibration with lens distortion from low-rank textures.
\newblock {\em CVPR 2011}, pages 2321--2328, 2011.

\end{thebibliography}
\bibliographystyle{plain}

\end{document}